\documentclass[11pt, letterpaper]{article}
\usepackage[utf8]{inputenc}
\usepackage[left=1in, right=1in]{geometry}
 
\usepackage{graphicx}

\usepackage{amssymb,amsmath}

\usepackage[authoryear, round]{natbib}

\usepackage[hidelinks]{hyperref}
\usepackage{booktabs}
\title{Efficient data-driven gap filling of satellite image time series using deep neural networks with partial convolutions}

\usepackage{authblk}
\author[1]{Marius Appel} % TODO: affiliation
\affil[1]{University of Münster, Insitute for Geoinformatics, Heisenbergstr. 2, 48149 Münster, Germany; Contact: \url{marius.appel@uni-muenster.de}}
\date{} % TODO

\begin{document}

\maketitle

\begin{abstract}
The abundance of gaps in satellite image time series often complicates the application of deep learning models such as convolutional neural networks for spatiotemporal modeling.
Based on previous work in computer vision on image inpainting, this paper shows how three-dimensional spatiotemporal partial convolutions can be used as layers in neural networks to fill gaps in satellite image time series. To evaluate the approach, we apply a U-Net-like model on incomplete image time series of quasi-global carbon monoxide observations from the Sentinel-5P satellite. Prediction errors were comparable to two considered statistical approaches while computation times for predictions were up to three orders of magnitude faster, making the approach applicable to process large amounts of satellite data. Partial convolutions can be added as layers to other types of neural networks, making it relatively easy to integrate with existing deep learning models.  However, the approach does not quantify prediction errors and further research is needed to understand and improve model transferability. The implementation of spatiotemporal partial convolutions and the U-Net-like model is available as open-source software.
\end{abstract}


\section{Introduction}
\label{sec:introduction}

Deep learning (DL) and particularly convolutional neural networks (CNNs) have been exceptionally successful for satellite image analysis tasks including object detection and segmentation. In recent years, DL models have been increasingly used also for modeling continuous spatiotemporal phenomena such as soil moisture \citep{elsaadani2021} or air temperature \citep{amato2020}.
Developments in \citet{debezenac2019} demonstrate that such models can reproduce fundamental physical properties of processes such as advection and diffusion in a purely data-driven way.  
\citet{rasp2021} even show that CNN-based models can perform medium range weather prediction with performance close to an operational physical model.	More generally, \citet{campsvalls2021} and \citet{reichstein2019} describe challenges and approaches for hybrid data-driven and physical modeling. However, a fundamental challenge when applying CNN-based spatiotemporal models on satellite image time series is the existence of missing values e.g, due to atmospheric conditions (in many cases clouds). 

Numerous approaches to fill gaps in spatiotemporal data have been proposed, including statistical methods based on discrete cosine transforms \citep{wang2012}, singular spectrum analysis  \citep{ghafarian2018, buttlar2014}, Markov random fields \citep{fischer2020}, spatiotemporal interpolation \citep{cressie2008, appel2020}, and more algorithmic methods such as quantile regression in local neighborhoods \citep{gerber2018}. In computer vision, a similar problem is referred to as \textit{image inpainting} or \textit{video inpainting}, where the aim is to restore corrupt parts of images and videos respectively. \citet{liu2018}  present a promising approach for the former, using a U-Net-like \citep{ronneberger2015} encoder / decoder model with \textit{partial convolutional layers} to fill gaps.  

Since it has been shown that CNN models are capable of modeling complex tasks and at the same time predictions can be computationally efficient, this study aims at (i) making CNNs with partial convolutions applicable to spatiotemporal Earth observation data from satellite image time series, (ii) evaluating prediction performance and computational aspects with regard to a quasi global atmospheric dataset, and (iii) discussing limitations, advantages, and future work towards purely data-driven modeling of spatiotemporal dynamics from incomplete datasets.

Notice that recently, \citet{dexing2022} similarly applied partial convolutions to a
spatiotemporal snow cover dataset. In contrast to their work, the presented approach uses 
three-dimensional partial convolutions, compares predictions to other methods, uses atmospheric
data on (quasi) global scale, and discusses the approach as a computationally efficient gap filling method.

The remainder of this paper is organized as follows. Section \ref{sec:methods} introduces the partial convolution operation and how it can be included in a model. Section \ref{sec:application} describes experimental details and the datasets used, before results are presented in Section \ref{sec:results}. A discussion of limitations of the approach and potential for future research is given in Section \ref{sec:discussion}, and Section \ref{sec:conclusions} concludes the paper.

\section{Methods}
\label{sec:methods}

\subsection{Partial convolutions}
\label{sec:partial-convolutions}

The following paragraph describes the partial convolution operation as introduced in \citet{liu2018}. 

Compared to ordinary convolutions, partial convolutions not only receive a data subset of the same size as the filter kernel but also a corresponding binary mask as input. Let $\mathbf{X}$, $\mathbf{M}$, and $\mathbf{K}$ be the data input, the mask input, and kernel weights respectively, all of identical shape. $\mathbf{M}$ has zeros for missing values and ones for valid observations, and we assume $\mathbf{X}$ is zero for missing values, too. We can then write the partial convolution operation as an ordinary convolution of $\mathbf{X}$ and $\mathbf{K}$ followed by multiplication with the number of elements in $\mathbf{K}$ divided by the number of ones in $\mathbf{M}$. The last step can be seen as applying a weight to adjust for missing values. Afterwards, the mask value of the center pixel is set to one if there is at least one valid observation in $\mathbf{X}$. Formally, applying a partial convolution at one location can be written as \citep{liu2018}	
\[
x' = 
\begin{cases}
	\mathbf{K} * (\mathbf{X} \odot \mathbf{M}) \frac{\sum \mathbf{1}}{\sum \mathbf{M}}& \textrm{if } \sum \mathbf{M} > 0 \\
	0 & \textrm{else,}
\end{cases}
\]
where $\odot$ is the element-wise multiplication, $*$ is the ordinary convolution, and $\mathbf{1}$ is an array with ones in the same shape as  $\mathbf{X}$, $\mathbf{K}$, and $\mathbf{M}$. The mask is updated by
\[
m' = 
\begin{cases}
	1 & \textrm{if } \sum \mathbf{M} > 0 \\
	0 & \textrm{else.}
\end{cases}
\]

Similar to ordinary convolutional layers in neural networks, a bias can be added and, if the input has multiple channels, separate convolutions (with different weights) are applied before computing per-pixel sums of the convolved channels. Notice that partial convolutions implicitly provide a padding strategy by simply extending the mask and data subset at the boundaries with zeros.

\subsection{Spatiotemporal models with partial convolutions}
\label{sec:spatiotemporal-models-with-partial-convolutions}

Similar to \citet{liu2018} and \citet{dexing2022}, we integrate partial
convolutional layers in a U-Net-like \citep{ronneberger2015} model
architecture consisting of encoder / decoder (or \emph{convolution} /
\emph{deconvolution}) parts and skip connections. The encoder reduces resolution by strided partial convolutions and (typically) increases the channel depth while
the decoder increases the resolution and combines lower resolution
output with output from associated layers of the encoder.

In contrast to \citet{liu2018} and \citet{dexing2022}, our model applies
\emph{three-dimensional} partial convolutions. The input is a
spatiotemporal block \(X\) of size \(n_x \times n_y \times n_t\) and a binary
mask \(M\) of the same size, where 1 represents that the corresponding
value in \(X\) is valid and 0 represents missing values.

At first, one or more partial convolutional blocks are applied to the
input. A block applies one or more partial convolutions sequentially
with a user-defined number of kernels, where the last convolution
applies striding. Each partial convolutional layer is followed by a leaky ReLU (\(\alpha = 0.1\)) activation function. Once the lowest spatiotemporal resolution is
reached, the output is upsampled again to increase spatiotemporal
resolution. The upsampled output is then concatenated with the output of
the associated block from the convolutional phase, before another
partial convolutional block (without striding) and leaky ReLU activation
is applied and finally the original size of the block is reached.

Gaps are filled during the encoder part, while individual partial
convolutions are applied. Using larger kernels and applying more and/or
larger striding result in a faster filling of gaps. The depth of the
model hence must be adapted to the size of the gaps to make sure that
all gaps become filled. 

\begin{figure}
	{\centering 
		\includegraphics[width=\linewidth]{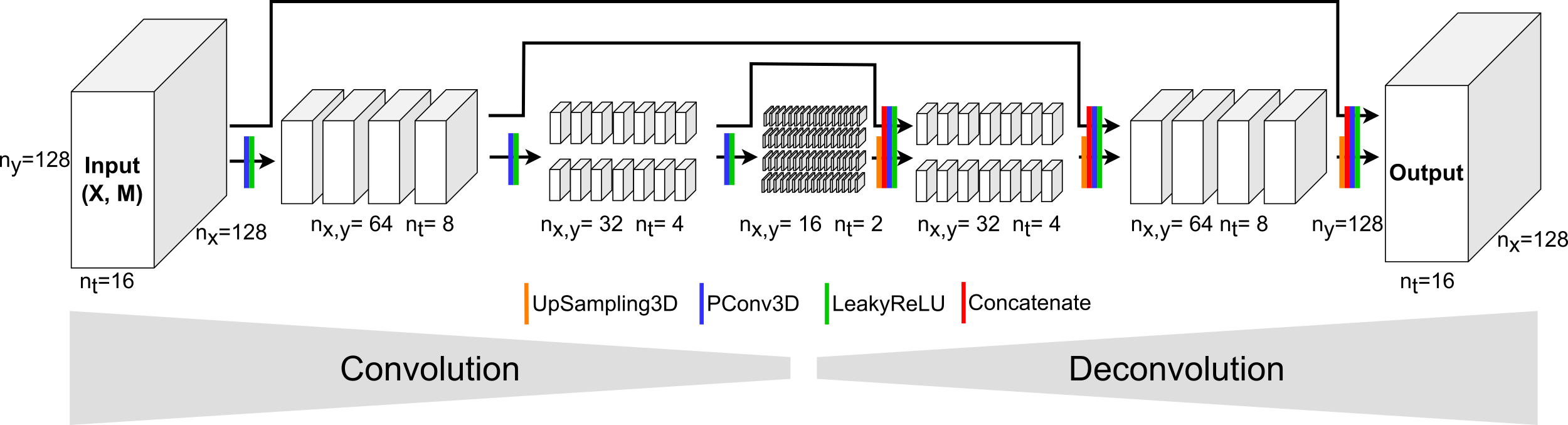} 
	}
	\caption{Architecture of our U-Net-like model (STpconv) with spatiotemporal partial convolutional layers. Notice that masks are omitted in the illustration but pass the network similarly to X.}
	\label{fig:model_architecture}
\end{figure}

Figure \ref{fig:model_architecture} illustrates the basic architecture of the model in an example, where spatiotemporal blocks have size \(128 \times 128 \times 16\),
spatiotemporal resolution is reduced by a striding factor of 2 in all
dimensions, there are three partial convolutional blocks in the encoder,
each consisting of a single partial convolutional layer, and the channel depth is
increased as the spatiotemporal resolution is decreased by using an increasing
number of filters.

The proposed model is highly customizable. Table \ref{tab:hyperparameters} lists
important hyperparameters with regard to the model architecture, training, and data preparation.

\begin{table}[t]
	\caption{Important hyperparameters related to the model architecture, model training, and data preparation.} 
	\label{tab:hyperparameters}
	\centering
	\small
	\begin{tabular}{p{5cm}p{3cm}p{6cm}}
		\toprule
		Model architecture & Model training & Data \\
		\midrule
		Number of convolutional blocks & Loss function &  Block-size  \\
		Convolutional layers per block & Optimizer & $n$   \\
		Spatiotemporal striding & Learning rate & Transformations,  trend / seasonality removal  \\
		Kernel sizes & Batch size & Method to add gaps  \\
		Number of filters & $\ldots$ &  $\ldots$ \\
		\bottomrule
	\end{tabular}
\end{table}

% \begin{table}[t]
	% \caption{Important hyperparameters related to the model architecture, model training, and data preparation.} 
	% \label{tab:hyperparameters}
	% \centering
	% \scriptsize
	% \setlength{\tabcolsep}{0.8mm}
	% \begin{tabular}{lll}
		%     \toprule
		%     Model architecture & Model training & Data \\
		%     \midrule
		%     \makecell[tl]{Number of convolutional blocks \\ Convolutional layers per block \\ Spatiotemporal striding \\ Kernel sizes \\ Number of filters} & \makecell[tl]{Loss function \\ Optimizer \\ Learning rate \\ Batch size \\ $\ldots$} & 
		%     \makecell[tl]{Block-size \\ \(n\) \\ Transformations \\ Trend / seasonality removal \\ Method to add gaps \\ $\ldots$} \\
		%     \bottomrule
		% \end{tabular}
	% \end{table}

Our implementation (Section \ref{sec:implementation}) allows to define architectural parameters differently per dimensions. For example, it is possible to add purely temporal or spatial partial convolutional blocks, and to define different kernel sizes in space and time. This can be used to optimize the model for specific spatiotemporal phenomena, depending on spatial and temporal resolution and autocorrelations.

\subsection{Addition of artificial gaps}
\label{sec:addition-of-artificial-gaps}

To assess prediction errors during model training and validation, data
must be predicted at locations with available measurements. Since the
aim is to predict (larger) gaps, the following strategy to add
artificial gaps to the input is applied. 

\begin{figure}
	\centering 
	\includegraphics[width=0.8 \linewidth]{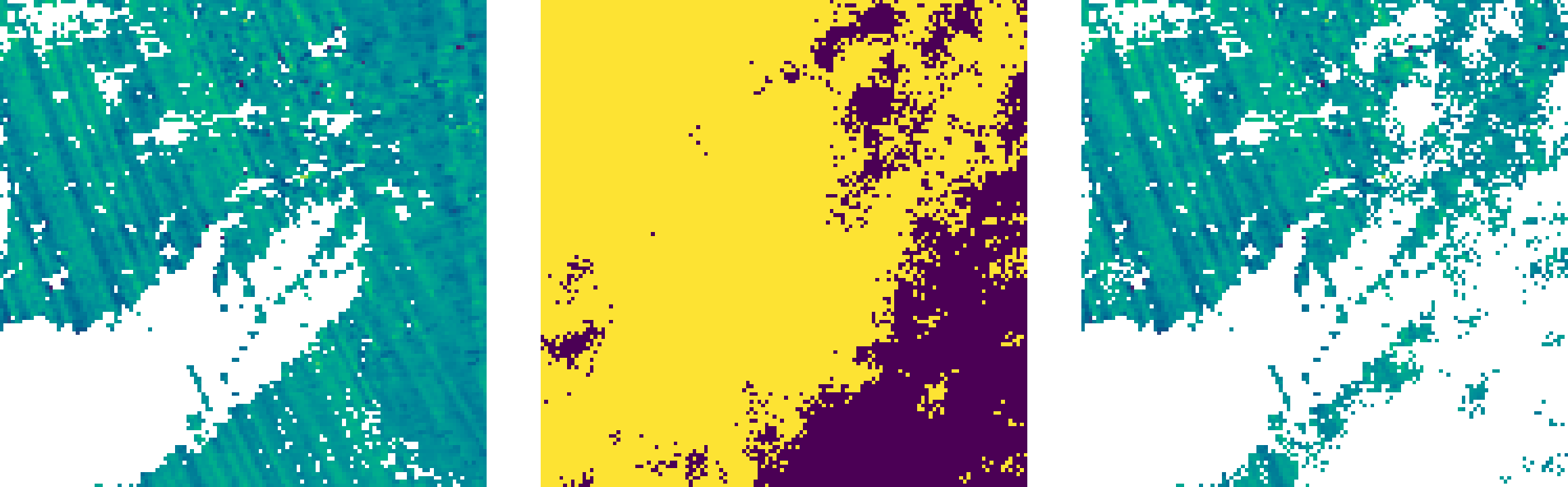} 
	\caption{Original image Y (left), additional mask (center) and masked image X (right) of a single time slice. X is used as input to the model. }\label{fig:masking}
\end{figure}

For each time slice in a spatiotemporal block \(Y\):	
\begin{enumerate}
	\item
	Simulate a two-dimensional Gaussian random field using a predefined
	covariance function \(\textrm{cov}(s, s')\) depending on the spatiotemporal distance between pairs of observations at spatiotemporal locations $s$ and $s'$ within a block. 
	\item
	Apply a threshold to select a moderate amount of data.
	\item
	Mask corresponding pixels from \(Y\) to create \(X\).
\end{enumerate}

\(X\) is then the input to our model and \(Y\) the target data used for
training and validation. The final amount of missing data in $X$ depends on both, the original amount of missing
values in $Y$ as well as on the amount of missing values in the synthetic mask. Since the former varies strongly among the data, the amount of missing values in the synthetic mask is less important than the actual shape of added masks. To allow the model to learn long range predictions of a process, it is important to generate artificial gaps of different sizes instead of just leaving out random pixels, resulting in only very small gaps. Figure \ref{fig:masking} illustrates how the additional mask is applied on a single time slice of a spatiotemporal block.

\subsection{Model training}
\label{sec:model-training}

We directly use the prediction error of values that have been omitted by the artificially added masks (Section \ref{sec:addition-of-artificial-gaps}) in our loss function and simply use the mean absolute error of observations that are available in $Y$ but not in $X$. 

Notice that it is of course possible to consider observations that are available in both $Y$ and $X$ and also the smoothness of transitions at gap boundaries \citep{dexing2022} in individual or in a combined loss function. Similarly, absolute errors can be replaced by squared errors to put more weight on outliers or extreme values. 
However, in the experiments (Section \ref{sec:experiments}), the choice and the design of the loss function had no clear effect of prediction performance and we therefore used the simple MAE metrics. 

We used the Adam optimizer \citep{adam} and an initial learning rate of 0.005 that is divided by 10 after every 10 epochs. Training runs for at least 30 epochs.

\subsection{Implementation}
\label{sec:implementation}

We implemented the partial convolutions in Python \citep{python} using TensorFlow \citep{tensorflow} with  Keras \citep{keras}. The implementation contains a class for three-dimensional partial convolutional layers and a class for the U-Net-like model that can be customized by hyperparameters (see Table \ref{tab:hyperparameters}). Notice that we reuse the original implementation of ordinary convolutional layers and an available open-source implementation of two-dimensional partial convolutions in Keras\footnote{see \url{https://github.com/MathiasGruber/PConv-Keras}}.

The source code of our implementation, including a small example dataset and a pre-trained model, is available on GitHub\footnote{\url{https://github.com/appelmar/STpconv}}.

\section{Application to Sentinel-5P data}
\label{sec:application}

To validate the proposed gap filling approach, we used imagery from
the European Sentinel-5P mission for
satellite-based monitoring of the atmosphere. The TROPOMI instrument on
board of the satellite measures atmospheric variables (total column observations of carbon monoxide, ozone, methane, nitrogen dioxide, and others\footnote{\url{http://www.tropomi.eu/data-products/level-2-products}}) at high
spatial resolution up to 3.5 km x 5.5 km and a revisit time of one day. Recently, Sentinel-5P  $\textrm{NO}_{\textrm{2}}$ observations have been integrated into operational Copernicus Atmosphere Monitoring Service forecasts by the European Centre for Medium-Range Weather Forecasts. 

For the experiments, we downloaded 4518 images of total column carbon
monoxide observations and corresponding per-pixel quality assessment
images of the offline processing stream from the Sentinel-5P Level 2
open data catalog on Amazon Web Services\footnote{see \url{https://registry.opendata.aws/sentinel5}}. Images have been recorded between 2021-01-01 and 2021-11-25.

\subsection{Data preprocessing}
\label{sec:data-preprocessing}

\begin{figure}	
	\centering
	\includegraphics[width=0.7\linewidth]{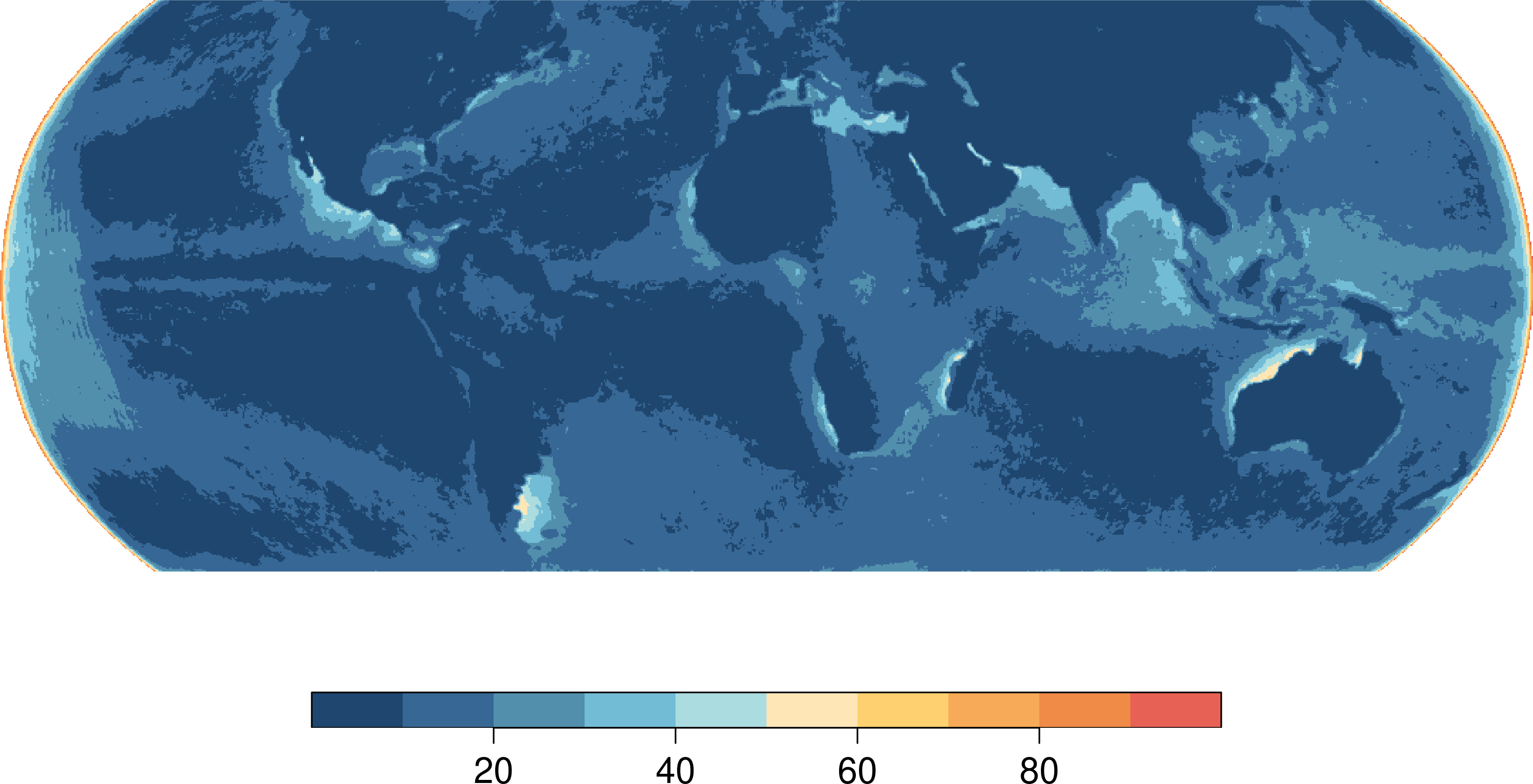} 
	\caption{Percentage of missing values per pixel time series after daily aggregation, cropping, and filtering by quality values (QA $>$ 0.5). Contains modified Copernicus Sentinel data [2021].}
	\label{fig:validobservations}
\end{figure}

We resampled the data to 0.1 degree spatial resolution, derived daily
aggregates, cropped to latitudes between -60 and 60 degrees, and,
following the official recommendation \citep{s5preadme}, only used
pixels with quality value larger than 0.5. Figure
\ref{fig:validobservations} shows the availability of valid observations
of the resulting dataset.

At most of the locations, less than 20\% of the days are missing in the prepared dataset. Some exceptions include the Northern coast of Australia, the Argentinian coast in the southern Atlantic ocean, Southeast Asia, and further coastal areas.

As a next step, we divided the data into spatiotemporal blocks of size
\(128 \times 128 \times 16\). We used a smaller block size in time, because 
gaps are typically larger in space. Furthermore, 
spatial autocorrelations between any two pixels are expected to show a longer range compared to 
temporal autocorrelations.

All of the preprocessing steps including
the resampling, aggregation, cropping, and filtering have been performed
in R \citep{R} using the \texttt{gdalcubes} package \citep{appel2019}.

\subsection{Experiments}
\label{sec:experiments}

To assess the performance of predictions, we randomly selected 500
spatiotemporal blocks for model training and applied two different
validation strategies on additional 250 randomly selected validation blocks:

\begin{enumerate}
	\def\labelenumi{\arabic{enumi}.}
	\item
	We added artificial gaps as for the training data (see Section \ref{sec:addition-of-artificial-gaps}), predict missing values, and computed prediction scores for predicted values that have been missing in $X$ but available in $Y$. 
	\item
	We used all available data but completely left out the last time slice of a block, which was then predicted and used to calculate prediction scores.
\end{enumerate}

In the following, the validation strategies are referred to as \emph{gap filling} and \emph{one-step ahead predictions} respectively. We used mean absolute error (MAE) and root mean squared error (RMSE) as metrics and
additionally calculated average computation times for predicting single
blocks. Error metrics refer only to pixels that have been additionally left out but do not include predictions of pixels that were available in both $X$ and $Y$. 

As a benchmark, we furthermore applied two naive methods
(simple block-wise mean prediction and linear time series
interpolation). The block-wise mean predictions simply use the empirical mean from all available observations of a block for all missing values whereas the time series interpolation independently interpolates time series of a block. We use the NumPy \citep{numpy} implementation in \texttt{numpy.interp}, with default parameters, meaning that if time series start or end with missing values, the first / last available observation is carried forward / backward. If a time series has no valid observations, it is not predicted and omitted in the calculation of prediction scores. 

As alternative data-driven approaches, we applied two statistical approaches on the same validation dataset. First, the method described in \citet{gerber2018}, which we refer to as \textit{gapfill}, constructs a spatiotemporal neighborhood for each missing value, whose size is increased until enough observations are available. Quantile regression is then applied on values in the neighborhood. This not only allows to predict the median but also to quantify prediction uncertainties as confidence intervals. For details, the reader is referred to the original publication in \citet{gerber2018}.

Second, we applied an efficient approximation of spatiotemporal Gaussian processes called multi-resolution approximations (referred to as \textit{stmra}) as proposed in \citet{appel2020}. This approach recursively partitions the area of interest into smaller regions, assuming conditional independence between different regions at the same partitioning level, and uses a basis function representation of processes within regions to approximate residuals from previous partitioning levels. Compared to traditional Geostatistical modeling, this approach can be used for large datasets but still requires fitting a spatiotemporal covariance model. For simplicity, we used a separable spatiotemporal covariance function where both functions use the exponential covariance model. Corresponding parameters have been fitted based on 10 randomly selected blocks.  

For computation time measurements, a single CPU core was used and the average after predicting all blocks from the validation set has been calculated. Computations have been executed on an Intel Core i7-7700HQ CPU with 16 GB main memory, and a 512 GB solid-state drive. Notice that our STpconv model has been trained on a NVIDIA A100 GPU with 40GB VRAM, before.

After trying out models with different complexities, different optimizers, and loss functions, a relatively small model with less than 50000 parameters has been selected for the experiment. This model turned out to offer a good compromise of prediction performance and computation times. Details of the model can be found in the Appendix (Table \ref{tab:modeldetails}).

\section{Results}
\label{sec:results}

\subsection{Prediction performance}
\label{sec:prediction-performance}

Table \ref{tab:results} presents results of the experiments for
different models and both validation strategies.

\begin{table}
	
	\caption{\label{tab:results}Prediction errors and computation times of different models and validation strategies. STpconv refers to the proposed model based on spatiotemporal partial convolutions.}
	\centering
	\begin{tabular}[t]{lrrrrr}
		\toprule
		\multicolumn{1}{c}{} & \multicolumn{2}{c}{Gap filling} & \multicolumn{2}{c}{One-step ahead forecasts} & \multicolumn{1}{c}{} \\
		\cmidrule(l{3pt}r{3pt}){2-3} \cmidrule(l{3pt}r{3pt}){4-5}
		& MAE & RMSE & MAE & RMSE & $t$ (s)\\
		\midrule
		Block-wise mean & 0.00274 & 0.00443 & 0.00295 & 0.00437 & 0.02\\
		Time series interp. & 0.00245 & 0.00412 & 0.00276 & 0.00421 & 0.51\\
		gapfill \citep{gerber2018} & 0.00166 & 0.00302 & NA & NA & 1480.23\\
		stmra \citep{appel2020} & 0.00155 & 0.00333 & 0.01062 & 0.01168 & 159.96\\
		STpconv & 0.00186 & 0.00321 & 0.00243 & 0.00357 & 0.44\\
		\bottomrule
	\end{tabular}
\end{table}

For \emph{gap filling}, the statistical and STpconv models outperformed
the naive approaches in terms of MAE and RMSE. Among the non-naive
methods, stmra performed best for MAE but worst in terms of RMSE,
suggesting that stmra had difficulties for predicting outliers or
extreme values. In general, gapfill seems to achieve slightly better
predictions than our partial convolutional neural network. However,
gapfill is not applicable to one-step ahead forecasting where a
complete time slice is missing because it cannot find a suitable
neighborhood. stmra performed significantly worse for one-step ahead
forecasting and our STpconv model clearly performs best.

Interestingly, predictions with the partial convolution model have been
even slightly faster than simple time series interpolation, and approximately 300 and 3000 times faster
compared to stmra and gapfill respectively.
However, notice that the computation time measurements include only the
prediction but not model fitting. Training the STpconv model on 500 spatiotemporal blocks took 20 minutes on a powerful GPU. In contrast, parameter optimization of the stmra approach took approximately 30 minutes on 10 randomly selected blocks using a CPU. However, the optimization time for stmra is generally hard to estimate in advance and can vary strongly for different starting values, covariance functions, and similar \citep{appel2020}. gapfill and the naive methods do not require any model fitting at all. 

\begin{figure}
	\centering
	\includegraphics[width=0.9 \linewidth]{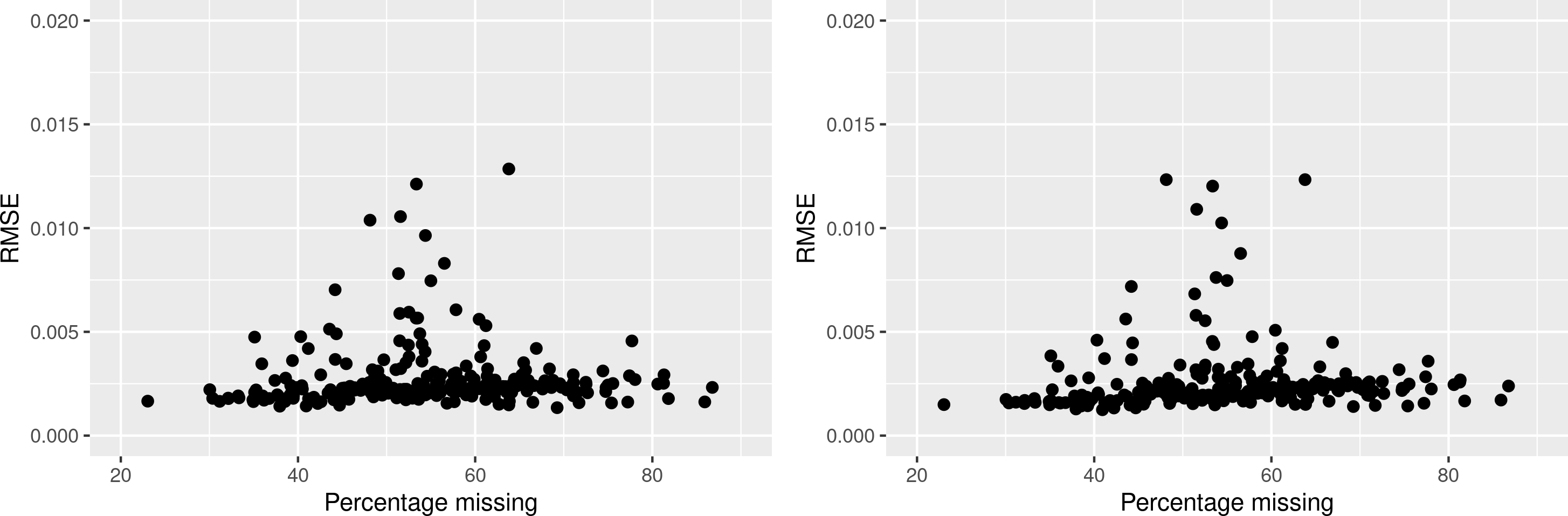}
	\caption{\label{fig:rmsebytiles}RMSE of individual spatiotemporal blocks
		by their percentage of missing values for predictions from STpconv (left)
		and gapfill (right). Each dot represents a block in the validation dataset and the x axis
		refers to the percentage of missing values in the input block (including added gaps).}
\end{figure}

Figure \ref{fig:rmsebytiles} additionally shows  
prediction errors of individual spatiotemporal blocks for gap filling plotted by their percentage of missing values.	
Here, we compare only our STpconv with gapfill but the results look very similar. 
Prediction errors seem relatively independent from the percentage of missing values even up to 90\% missing data. Interestingly, outliers with larger prediction errors refer to the same tiles for both approaches, though their RMSE values of course differs.

\subsection{Visual comparison}
\label{sec:visual-comparison}

\begin{figure}
	\centering
	\includegraphics[width=0.7 \linewidth]{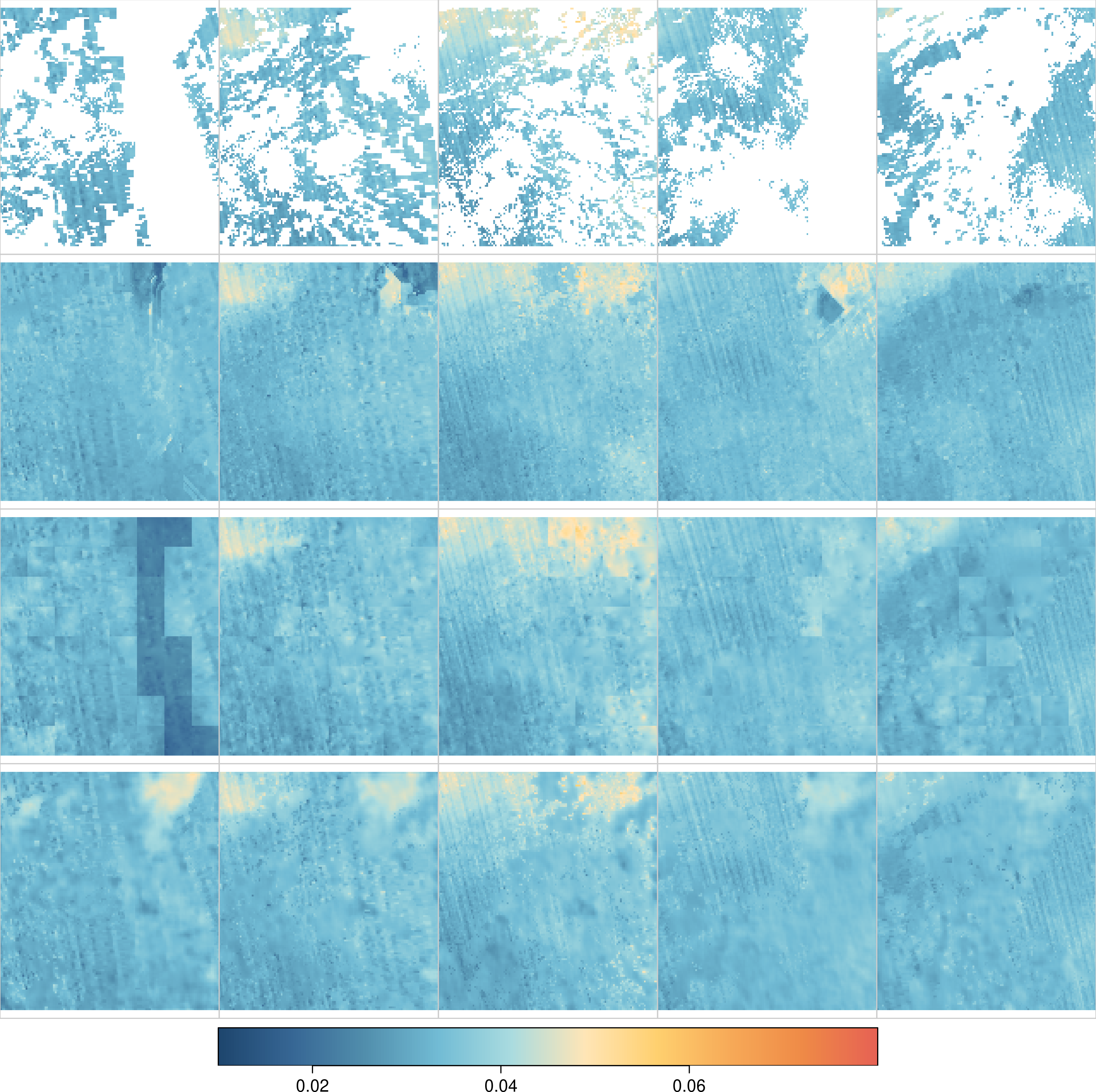}
	\caption{\label{fig:viualcomparison}Input and predictions of (left to right) five time slices using (from top to bottom) gapfill, stmra, and STpconv. Contains modified Copernicus Sentinel data [2021].}
\end{figure}

Figure \ref{fig:viualcomparison} shows input and predictions of five
example time slices for the two statistical models and the model based
on partial convolutions.

Predictions from the gapfill approach show relatively fine details,
mostly including the typical vertical stripe pattern. Due to the
selection of local neighborhoods for filling individual pixels, there
are some sharp edges. The stmra approach shows visible artefacts due to
the recursive partitioning of the area of interest, although the corresponding MAE was
best for gap filling. A model averaging using shifted partitioning grids
was not performed but would reduce artefacts \citep{appel2020}.
Predictions from the three-dimensional partial convolutions are
relatively smooth, such that the vertical stripe pattern is less visible
in predicted areas.

\subsection{CO mapping}
\label{sec:co-mapping}

\begin{figure}
	\centering 
	\includegraphics[width=0.7 \linewidth]{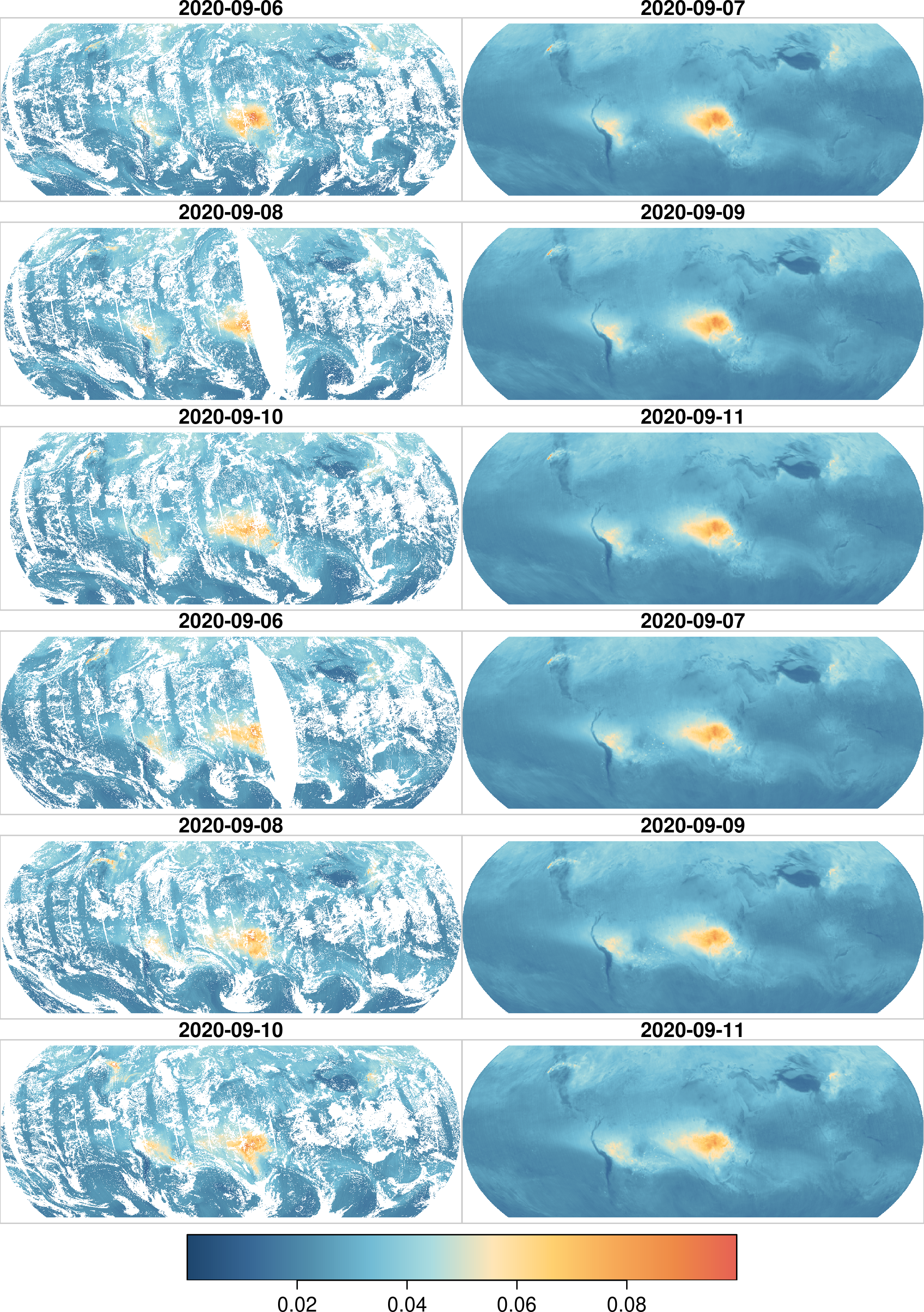} 
	\caption{Global daily original dataset (left) and model predictions (right) at 6 consecutive days. Contains modified Copernicus Sentinel data [2021].} 
	\label{fig:comap}
\end{figure}

We used the trained model to fill gaps in the original dataset and
produce quasi-global (60°S -- 60°N) daily maps of total column CO for
the time range of the data. Figure \ref{fig:comap} shows a few days of the original
incomplete and the corresponding filled images. Animated predictions can be
found in the supplementary material of this paper\footnote{for now, available at \url{http://appel.staff.ifgi.de/S5P_CO_demo.html}}. To reduce
artefacts at block boundaries, an overlap has been applied. For each
spatiotemporal block, only inner observations (leaving out a few pixels
at each side) are taken as predictions while adjacent blocks overlap by
two-times the number of left-out pixels.

Notice that Figure \ref{fig:comap} shows complete predictions, i.e. the model output
without merging with true observations, where available. The results look relatively smooth and do
not show any obvious block artefacts. The general dynamics of the process
seems well preserved. As expected, extremes tend to be underestimated.
Interestingly, the gap filled animation reveals some clearly visible
pattern of extreme values above the south-west Atlanic Ocean, mostly
close to the south American coastline. These extreme values up to 0.7 $\textrm{mol}~\textrm{m}^{-2}$
also occur in the original data but are only hardly visible due to the gaps.

\section{Discussion}
\label{sec:discussion}

Neural networks with spatiotemporal partial convolutional layers turned out to be able to efficiently fill gaps in atmopsheric (Sentinel-5P) image time series. While prediction errors in the experiments were similar to complex statistical models, computation times for prediction even outperformed naive time series interpolation. 
For model training, a powerful GPU is required although training times have been reasonable. Such GPUs are also available on cloud computing platforms for a few USD per hour\footnote{see, e.g. \url{https://aws.amazon.com/de/ec2/instance-types/p3} (accessed 2022-05-20)}, resulting in less than 3 USD total costs for training in our case.

However, compared to the naive and statistical approaches, models based on partial convolutions come with the risk of excessive prediction errors under extrapolation conditions i.e., when the neural network is applied to data dissimilar to the input data in the training set \citep{meyer2021}, or if there is hardly any observation available in a block. More complex loss functions could improve the robustness of predictions. For example, terms that penalize if block-wise means of predictions deviate too much from actual means of available pixels could be added. The approach also does not provide uncertainty estimates such as prediction intervals as the statistical models do. \citet{zammitmangion2020} therefore integrate a CNN for modeling spatiotemporal dynamics into a hierarchical statistical framework.   

\begin{figure}
	{
		\centering \includegraphics[width=0.9\linewidth]{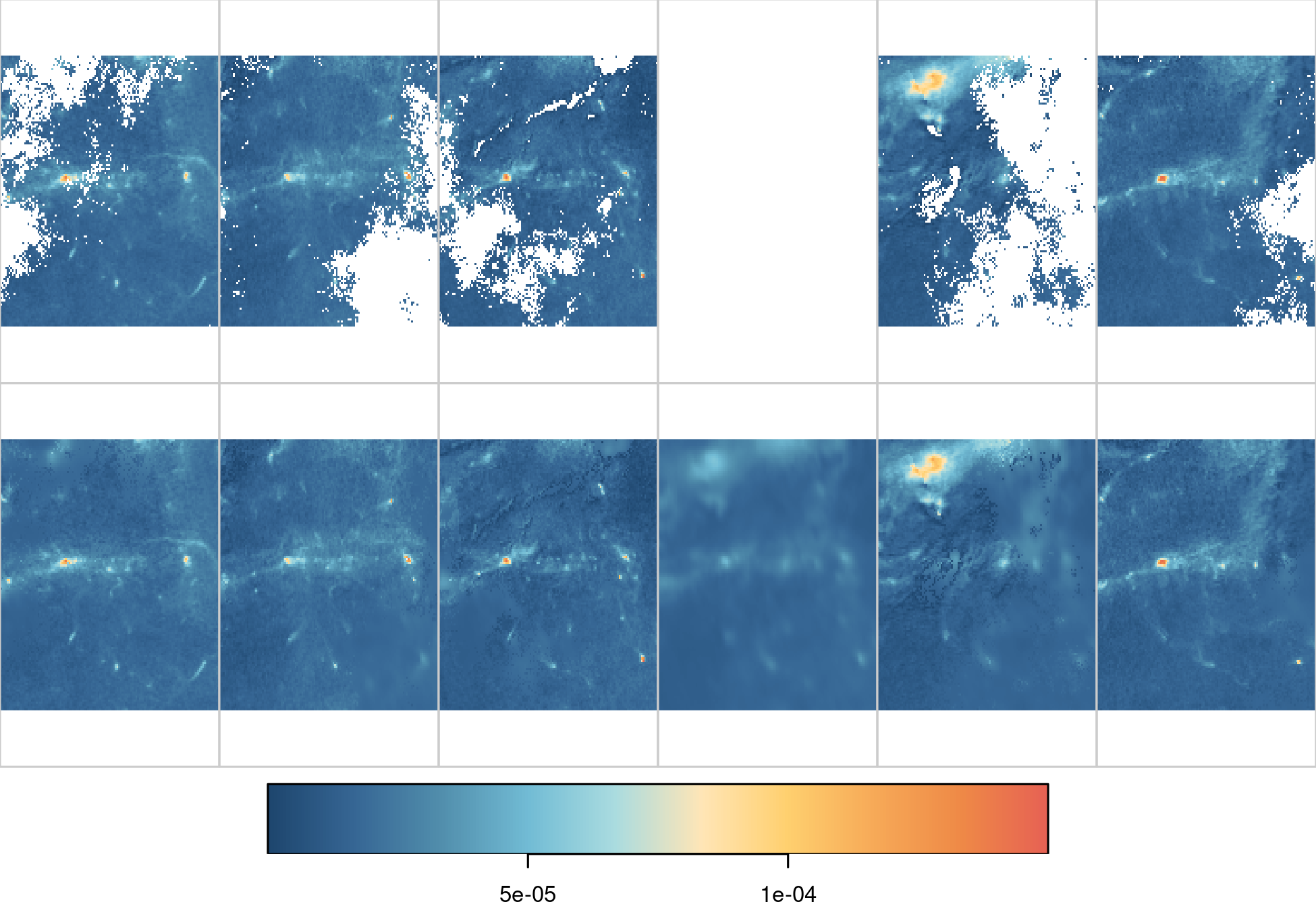} 
		\caption{\label{fig:no2}Results after applying a model that has been trained on CO data to 
			$\textrm{NO}_{\textrm{2}}$ observations. The top row shows six original $\textrm{NO}_{\textrm{2}}$ time slices, where one slice is completely missing, and the bottom row shows gaps filled with corresponding predictions. Contains modified Copernicus Sentinel data [2021].}\label{fig:unnamed-chunk-8}	
	}
	
\end{figure}

This study has not investigated how well models based on spatiotemporal partial convolutions work for other types of data such as optical imagery, or land surface temperatures. \citet{dexing2022} suggest that a similar method based on partial convolutions seems to work well for snow cover mapping, too. Similarly, it would be interesting to study how models trained on one variable can be used to predict other variables. Figure \ref{fig:no2} exemplarily shows results when the model trained on CO observations is applied to $\textrm{NO}_{\textrm{2}}$ data, after scaling to a similar value
range. However, further experiments are needed to evaluate
model transferability among other Sentinel-5P variables. To improve predictions, it might also be important to add other Sentinel-5P variables and/or further external variables such as elevation to the model. 

Given the amount of hyperparameters, tuning can be time consuming and make the optimization of the model architecture, data preprocessing, and similar to specific datasets difficult in practical applications. In the presented experiments, we have found that data-related parameters such as the spatiotemporal block size had a stronger effect on prediction performance compared to architectural parameters such as the number of filters per layer. However, this might me different when using much larger sets of training data. Further, it would be possible to use partial convolutional layers in other model architectures than the presented U-Net-like model. For example, it would be interesting to explore models
following the idea of residual neural networks \citep{he2016}, where data at different spatial and/or temporal resolutions is provided as input to the model and residuals from filling gaps at lower resolution are recursively considered by partial convolutional blocks.  

DL based models have become promising for modeling the dynamics of continuous spatiotemporal phenomena \citep{debezenac2019,zammitmangion2020,rasp2021, keisler2022}. Since the abundance of gaps in satellite-derived observations makes their application often difficult, it would be very interesting to integrate partial convolutions into similar models applied on incomplete data. As a first experiment, one could replace convolutional layers in \citet{debezenac2019} with partial convolutional layers and study how the performance of forecasts changes with the amount of missing values.

\section{Conclusions}
\label{sec:conclusions}

This paper discussed the use of neural networks based on partial convolutional layers for efficiently filling gaps in satellite image time series. Prediction results have been comparable to statistical methods while computations of predictions were between two and three orders of magnitude faster. However, future research should be conducted to assess model transferability and prediction errors, and to combine the approach with deep learning models of spatiotemporal dynamics in Earth observation data.

\section*{Acknowledgments}

This research has been funded by the Deutsche Forschungsgemeinschaft (DFG, German Research Foundation) – 396611854. 
Thanks to Edzer Pebesma for comments and suggestions to improve the final manuscript.

\section*{Data availability statement}

Data used for model training and validation have been made available on Zenodo under the DOI \texttt{10.5281/zenodo.6838651} \citep{appel_marius_2022_6838652}.
The dataset also includes predictions from other models discussed in Section \ref{sec:experiments}.

To reproduce results, the method has been made available as open-source software at GitHub\footnote{\url{https://github.com/appelmar/STpconv}}.

\section*{Appendix}

\subsection*{Model details}
\label{sec:model-details}

\begin{table}[h!]
	\caption{Selected hyperparameters of the used model with partial convolutions.}
	\label{tab:modeldetails}
	\centering
	\begin{tabular}{ll} \toprule
		Parameter & Value \\
		\midrule
		Number of partial convolutional blocks & 2 \\
		Partial convolutional layers per block & 1 \\
		Spatiotemporal striding for all blocks & [2, 2, 2] \\
		Kernel sizes & [3, 3, 3] \\
		Number of filters per block & [16, 16] \\
		Loss function & MAE (on gap pixels only) \\
		Optimizer & Adam \\
		Learning rate & adaptive (initially 0.005)\\
		Batch size & 6 \\
		Data block size & [128, 128, 16] \\
		Transformations & None \\
		\bottomrule
	\end{tabular}
\end{table}

\clearpage
\bibliographystyle{apalike}
\bibliography{references}

%\printbibliography

\end{document}